\documentclass{article} 
\usepackage{times}
\usepackage{amsfonts}

\usepackage{graphicx} 
\usepackage{subfigure}

\usepackage{natbib}

\newif\ifenoughspace
\enoughspacefalse

\title{Better Mixing via Deep Representations}

\author{
Yoshua Bengio, Gr\'egoire Mesnil, Yann Dauphin and Salah Rifai\\
Department of Computer Science and Operations Research\\
University of Montreal\\
Montreal, H3C 3J7 \\
}

\newcommand{\sigm}{\mathop{\mathrm{sigmoid}}}

\begin{document}

\maketitle

\begin{abstract}
  It has previously been hypothesized, and supported with some experimental evidence, that deeper representations, when well trained, tend to do a better job at disentangling the underlying factors of variation.  We study the following related conjecture: better representations, in the sense of better disentangling, can be exploited to produce faster-mixing Markov chains. Consequently, mixing would be more efficient at higher levels of representation.  To better understand why and how this is happening, we propose a secondary conjecture: the higher-level samples fill more uniformly the space they occupy and the high-density manifolds tend to unfold when represented at higher levels.  The paper discusses these hypotheses and tests them experimentally through visualization and measurements of mixing and interpolating between samples.
\end{abstract}

\section{Introduction and Background}

Deep learning algorithms attempt to discover multiple levels of
representation of the given data (see~\citep{Bengio-2009} for a review),
with higher levels of representation defined hierarchically in terms of
lower level ones. The central motivation is that higher-level
representations can potentially capture higher-level abstractions relevant
to the distribution of interest. Mathematical results in the case of
specific function families have shown that choosing a sufficient depth of
representation can yield exponential benefits, in terms of
size of the model, to represent some
functions~\citep{Hastad86-small,Hastad91,Bengio-localfailure-NIPS-2006-small,Bengio+Lecun-chapter2007,Bengio+Delalleau-ALT-2011-short}. The intuition
behind these theoretical advantages is that lower-level features or latent
variables can be {\em re-used} in many ways to construct higher-level ones,
and the potential gain becomes exponential with respect to depth of the
circuit that relates lower-level features and higher-level ones (thanks to
the exponential number of paths in between). The ability of deep learning
algorithms to construct abstract features or latent variables on top of the
observed random variables therefore relies on this idea of re-use, which
brings with it not only computational but also statistical advantages in
terms of statistical power, e.g., as already exploited in prior machine
learning work such as multi-task learning
algorithms~\citep{caruana95-small,baxter97,CollobertR2008-small} and
learning algorithms involving {\em parameter
  sharing}~\citep{Lang+Hinton88,LeCun89a}.

There is another -- less commonly discussed -- motivation for deep
representations, introduced in~\citet{Bengio-2009}: the idea that they may,
to some extent, help to {\em disentangle} the underlying factors of
variation. Clearly, if we had learning algorithms that could do a good job
of discovering and separating out the underlying causes and factors of
variation present in the data, it would make further processing (typically,
taking decisions) much easier. One could even say that the ultimate goal of
AI research is to build machines that can understand the world around us,
i.e., disentangle the factors and causes it involves,
so progress in that direction seems important. If learned representations
do a good job of disentangling the underlying factors of variation, it is
clear that learning (on top of these representations, e.g., towards
specific tasks of interest) becomes substantially easier because disentangling
counters the effects of the curse of dimensionality. In fact, in the
extreme case, and where we also observe the target variables and effects of
decisions, there is no need for further learning, only good inference.
Several observations have been made and reported that suggest that some deep learning
algorithms indeed help to disentangle the underlying factors of 
variation~\citep{Goodfellow2009-short,Glorot+al-ICML-2011-small}.
However, it is not clear why, and to what extent in general (if any),
different deep learning algorithms may sometimes help this disentangling.

Most deep learning algorithms are based on some form of unsupervised
learning, hence capturing salient structure in the data
distribution. Whereas deep learning algorithms have mostly been used to
learn features and exploit them for classification or regression tasks,
their unsupervised nature also means that in several cases they can be used
to generate samples. In general the associated sampling algorithms involve
a Markov Chain and MCMC techniques, and these can notoriously suffer from a
fundamental problem of {\em mixing}: it is difficult for the Markov chain
to jump from one mode of the distribution to another, when these are
separated by large low-density regions, a common situation in real-world
data, and under the {\em manifold hypothesis}~\citep{Cayton-2005,Narayanan+Mitter-NIPS2010-short}. 
This hypothesis states that
natural classes present in the data are associated with low-dimensional
regions in input space (manifolds) near which the distribution
concentrates, and that different class manifolds are well-separated by
regions of very low density. Slow mixing means that consecutive samples
tend to be correlated (belong to the same mode) and that it takes many
consecutive sampling steps to go from one mode to another and even more to
cover all of them, i.e., to obtain a large enough representative set of samples
(e.g. to compute an expected value under the target distribution).
 This happens because these jumps through the empty low-density void
between modes are unlikely and rare events. When a learner has a poor model
of the data, e.g., in the initial stages of learning, the model tends to
correspond to a smoother and higher-entropy (closer to uniform) distribution, putting
mass in larger volumes of input space, and in particular, between the modes
(or manifolds). This can be visualized in generated samples of images, that
look more blurred and noisy. Mixing is therefore initially easy for such poor models.
However, as the model improves and its corresponding distribution sharpens
near where the data concentrate, mixing becomes considerably slower. Since
sampling is an integral part of many learning algorithms (e.g., to estimate
the log-likelihood gradient), slower mixing then means slower or poorer
learning, and one may even suspect that learning stalls at some point
because of the limitations of the sampling algorithm.  To improve mixing, a powerful idea that
has been explored recently for deep learning 
algorithms~\citep{Desjardins+al-2010-small,Cho10IJCNN-small,Salakhutdinov-2010-small,Salakhutdinov-ICML2010-small} 
is {\em tempering}~\citep{Neal94b}.
The idea is use smoother densities (associated with {\em higher temperature}
in a Boltzmann Machine or 
Markov Random Field formulation) to make quick but approximate jumps 
between modes, but
use the sharp ``correct'' model to actually generate the samples of interest
inside these modes, and allow samples to be exchanged between the different
levels of temperature.

Here we want to discuss another possibly related idea, and claim that
{\em mixing is easier when sampling at the higher levels of
  representation}.  The objective is not to propose a new sampling
algorithm or a new learning algorithm, but rather to investigate this
hypothesis through experiments using existing deep learning algorithms. The
objective is to further our understanding of this hypothesis through more
specific hypotheses aiming to explain why this would happen, using further
experimental validation to test these more specific hypotheses. The idea
that deeper generative models produce not only better features for
classification but also better quality samples (in the sense of better
corresponding to the target distribution being learned) is not novel and
several observations support this hypothesis
already, some quantitatively~\citep{Salakhutdinov+Hinton-2009}, 
some more qualitative~\citep{Hinton06-small}. The specific contributions
of this paper is to focus on why the samples may be better, and in
particular, why the chains may converge faster, based on the previously
introduced idea that deeper representations can do a better job
of disentangling the underlying factors of representation.

\section{Hypotheses}

We first clarify the first hypothesis to be tested here.
\begin{center}
\framebox{
\begin{minipage}{0.8\linewidth}
{\bf Hypothesis H1: Depth vs Better Mixing.} A successfully trained deeper architecture
has the potential to yield representation spaces in which Markov chains mix faster.
\end{minipage}
}\\
\end{center}

If experiments validate that hypothesis, the most important next question is: why?
The main explanation we conjecture is formalized in the following hypothesis.
\begin{center}
\framebox{
\begin{minipage}{0.8\linewidth}
{\bf Hypothesis H2: Depth vs Disentangling.} Part of the explanation of {\bf H1} is that
deeper representations can better disentangle the underlying factors of variation.
\end{minipage}
}\\
\end{center}
Why would that help to explain {\bf H1}? Imagine an abstract (high-level)
representation for object image data in which one of the factors is the
``reverse video bit'', which inverts black and white, e.g., flipping that
bit replaces intensity $x \in [0,1]$ by $1-x$. With the default value of 0,
the foreground object is dark and the background light.  Clearly, flipping
that bit does not change most of the other semantic characteristics of the
image, which could be represented in other high-level features. However,
for every image-level mode, there would be a reverse-video counterpart mode
in which that bit is flipped: these two modes would be separated by
vast ``empty'' (low density) regions in input space, making it very
unlikely for any Markov chain in input space (e.g. Gibbs sampling in an
RBM) to jump from one of these two modes to another, because that would require
most of the input pixels or hidden units of the RBM to simultaneously
flip their value. Instead, if we consider the high-level representation 
which has a ``reverse video''
bit, flipping only that bit would be a very likely event 
under most Markov chain transition probabilities,
since that flip would be a small change preserving high probability.
As another example, imagine that some of the bits of the high-level
representation identify the category of the object in the image, independently of pose,
illumination, background, etc. Then simply flipping one of these
object-class bits would also drastically change the raw pixel-space image,
while keeping likelihood high. Jumping from an object-class mode to another
would therefore be easy with a Markov chain in representation-space, 
whereas it would be much less likely in raw pixel-space.

Another point worth discussing (and which should be considered in future
work) in {\bf H2} is the notion of {\em degree} of disentangling.  Although
it is somewhat clear what a completely disentangled representation would
look like, deep learning algorithms are unlikely to do a perfect job of
disentangling, and current algorithms do it in stages, with more abstract
features being extracted at higher levels. 
Better disentangling would mean that {\em some} of the learned features
have a higher mutual information with {\em some} of the known factors. One would
expect at the same time that the features that are highly predictive of one
factor be less so of other factors, i.e., that they specialize to one or a
few of the factors, becoming {\em invariant} to others. Please note here
the difference between the objective of learning disentangled
representations and the objective of learning invariant features (i.e.,
invariant to some specific factors of variation which are considered to be
like nuisance parameters). In the latter case, one has to know ahead of
time what the nuisance factors are (what is noise and what is signal?).
In the former, it is not needed: we only seek to separate out the factors
from each other. Some features should be sensitive to one factor and
invariant to the others. 

Let us now consider additional hypotheses that specialize {\bf H2}.
\begin{center}
\framebox{
\begin{minipage}{0.8\linewidth}
{\bf Hypothesis H3: Disentangling Unfolds and Expands.} Part of the explanation of {\bf H2} is that
more disentangled representations tend to {\bf (a)} 
unfold the manifolds near which raw data concentrates, as well as
{\bf (b)} expand the relative volume occupied by high-probability points near
these manifolds.
\end{minipage}
}\\
\end{center} {\bf H3(a)} says is that disentangling has the effect that
the projection of high-density manifolds in the high-level representation space are smoother
and easier to model than the corresponding high-density manifolds in raw
input space. Let us again use an object recognition analogy. If we have
perfectly disentangled object identity, pose and illumination, the
high-density manifold associated with the
distribution of features in high-level representation-space is flat: we can
make large moves in that space (e.g., completely change the lighting) and
yet stay in a high-probability region. In fact we can make any move inside
some bounds and convex constraints and stay in a high-probability region,
so that the distribution of high-level features may look locally more
uniform, which is a consequence of {\bf H3(b)}. A good high-level
representation does not need to allocate as much real estate (sets of
values) for unlikely configurations. This is already what most unsupervised
learning algorithms tend to do. For example, dimensionality reduction
methods such as the PCA tend to define representations where most
configurations are likely (but these only occupy a subspace of the possible
raw-space configurations).  Also, in clustering algorithms such as k-means,
the training criterion is best minimized when clusters are approximately
equally-weighted, i.e., the average posterior distribution over cluster
identity is approximately uniform. Something similar is observed in the
brain where different areas of somatosensory cortex correspond to different
body parts, and the size of these areas adaptively depends~\citep{Flor-2003} 
on usage of these (i.e.,
more frequent events are represented more finely and less frequent ones are
represented more coarsely).  Again, keep in mind that the actual
representations learned by deep learning algorithms are not perfect, but
what we will be looking for here is whether deeper representations
correspond to more unfolded manifolds and to more locally uniform
distributions, with high-probability points occupying an overall greater
volume (compared to the available volume).

\section{Representation-Learning Algorithms}

The learning algorithms used in this paper to explore the preceding hypotheses are
the Deep Belief Network or DBN~\citep{Hinton06-small}, trained by stacking
Restricted Boltzmann Machines or RBMs, and the Contractive Auto-Encoder or
CAE~\citep{Rifai+al-2011-small}, for which a sampling algorithm was
recently proposed~\citep{Rifai-icml2012}.

See~\citet{Bengio-2009} for a detailed review of RBMs and DBNs. Each layer
of the DBN is trained as an RBM, and a 1-layer DBN is just an RBM. An RBM
defines a joint distribution between a hidden layer $h$ and a visible layer $v$. 
Gibbs sampling at the top level of the DBN is used to obtain samples from the
model: the sampled top-level representations are stochastically projected
down to lower levels through the conditional distributions $P(v|h)$
defined in each RBM. To avoid unnecessary additional noise, and like
previous authors have done, at the last stage of this process (i.e.
to obtain the raw-input level samples), only the mean-field values
of the visible units are used, i.e., $E[v|h]$. In the experiments on face
data (where grey levels matter a lot), a Gaussian RBM is used at the
lowest level. 

An auto-encoder~\citep{Lecun-these87,hinton1994amd-small} is parametrized
through an encoder function $f$ mapping an input-space vector $x$ to a
representation-space vector $h$, and a decoder function $g$ mapping a
representation-space vector $h$ to an input-space reconstruction $r$. The
experiments with the CAE are with $h=f(x)=\sigm(W x + b)$ and
$r=g(h)=\sigm(W^T h + c)$.  The CAE is a regularized auto-encoder, with
tied weights (input to hidden weights are the transpose of hidden to
reconstruction weights). The CAE is trained to minimize a reconstruction
loss (cross-entropy here) plus a contractive regularization penalty
$\alpha ||\frac{\partial f(x)}{\partial x}||^2_F$ (which is the sum of the
elements of the Jacobian matrix). Like RBMs, CAE layers can be stacked to
form deeper models, and one can either view them as deep auto-encoders (by
composing the encoders together and the decoders together) or like in a
DBN, as a top-level model (from which one can sample) coupled with encoding
and decoding functions into the top level (by composing the lower-level
encoders and decoders).  A sampling algorithm was recently proposed for
CAEs~\citep{Rifai-icml2012}.  The idea is to alternate between projecting
through the auto-encoder (i.e. performing a reconstruction) and adding
Gaussian noise $J J^T \epsilon$ in the directions of variation captured by the auto-encoder
(in the Jacobian matrix $J=\frac{\partial f(x)}{\partial x}$ of the encoder function). 

\section{Experiments}

The experiments have been performed on the MNIST digits
dataset~\citep{LeCun98-small} and the Toronto Face
Database~\citep{Susskind2010}, TFD. The former has been heavily used to evaluate
many deep learning algorithms, while the latter is interesting because of
the manifold structure it displays, and for which control factors (such as
emotion and person identity) are known.

The DBNs tested on MNIST have 768-1024-1024 layer sizes (28$\times$28 input),
and 2304-512-1024 on TFD (48$\times$48 input). The CAEs have sizes 768-1000-1000
and 2304-1000-1000 on MNIST and TFD respectively.

\subsection{Sampling at Different Depths}

\subsubsection{Better Samples at Higher Levels}

\begin{figure*}
\begin{center}
    \includegraphics[width=0.95\textwidth]{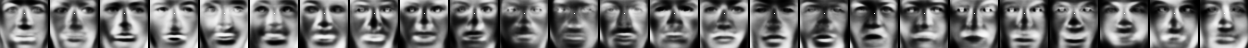}
    \includegraphics[width=0.95\textwidth]{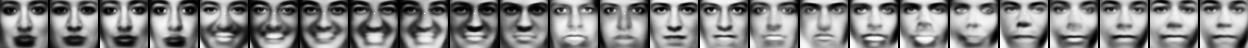}
    \includegraphics[width=0.95\textwidth]{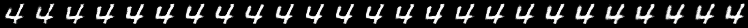}
    \includegraphics[width=0.95\textwidth]{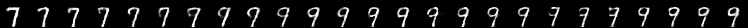}
\end{center}

\caption{Sequences of $25$ samples generated with a CAE
on TFD (rows 1 and 2) and with an RBM on MNIST (rows 3 and 4). 
{\small On TFD, the second layer clearly allows to get
quickly from woman samples (left) to man samples (right) passing by various facial
expressions whereas the first layer shows poor samples.
Bottom rows: On MNIST, the first layer gets stuck
in the same sample while the second layer allows to mix among classes.}  }

\label{fig:seq}
\end{figure*}

To test {\bf H1}, we first plot sequences of samples at various depths. One can
verify in Fig.~\ref{fig:seq} that samples obtained at deeper layers are
visually more likely and mix faster.

In addition, we measure the quality of the obtained samples, using a
procedure for the comparison of sample generators described in
\citet{Breuleux+Bengio-2011}. It measures the log-likelihood of a test set
under the density computed from a Parzen window density estimator built on
$10,000$ generated samples. Log-likelihoods for different models are presented
in Table~\ref{tab:res-knn-svm} (rightmost columns).  Those results also suggest that the
quality of the samples is higher if the Markov chain process used for sampling
takes place in the upper layers.

This observation agrees with {\bf H3(b)} that moving in higher-level representation
spaces where the manifold has been expanded provides higher quality samples than
moving in the raw input space where it may be hard to stay in high density
regions.

\subsubsection{Visualizing Representation-Space by Interpolating Between Neighbors}

\begin{figure*}
\begin{center}
\subfigure[Interpolating between an example and its $200$-th nearest neighbor]
{
    \includegraphics[width=0.24\textwidth]{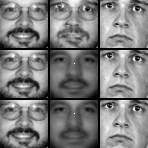}
    \includegraphics[width=0.24\textwidth]{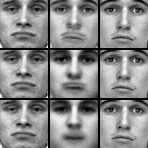}
    \includegraphics[width=0.24\textwidth]{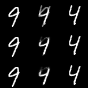}
    \includegraphics[width=0.24\textwidth]{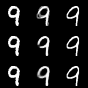}
    }
\subfigure[Interpolating between an example and its nearest neighbor]
{
    \includegraphics[width=0.24\textwidth]{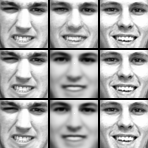}
    \includegraphics[width=0.24\textwidth]{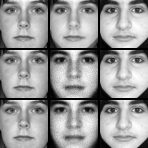}
    \includegraphics[width=0.24\textwidth]{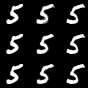}
    \includegraphics[width=0.24\textwidth]{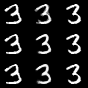}
    }

\subfigure[Sequences of points interpolated at different depths]
{
    \includegraphics[width=0.96\textwidth]{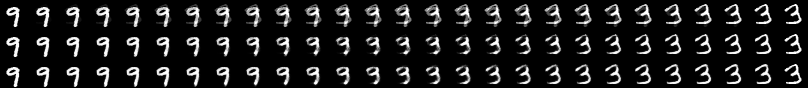}
    \label{fig:seqinterpol}
}
\end{center}

\caption{ Linear interpolation between a data sample and the 200-th (a)
and 1st (a) nearest neighbor, at various depths (top row=input space, middle row=1st layer,
bottom row=2nd layer).  In each $3\times3$ block the left and right columns are
test examples while the middle column is the interpolated point's input image.
Interpolating at higher levels clearly gives more likely samples. Especially in
the raw input space (a, 2nd block), one can see two mouths overlapping while only one mouth appears
for the interpolated point at the 2nd layer. Interpolating with the 1-nearest neighbor
does not show any difference between the levels. In (c), we interpolate between samples
of different classes, at different depths (top=raw input, middle=1st layer, bottom=2nd layer).
Note how in lower levels one has to go through unplausible patterns, whereas in the deeper
layers one almost jumps from a high-density region (of one class) to another.}
\label{fig:interpol}
\end{figure*}

According to {\bf H3(a)}, deeper layers tend to locally unfold the manifold
near high-densities regions of the input space, while according to {\bf H3(b)}
there should be more relative volume taken by plausible configurations in representation-space.
Both of these would imply that convex combinations of neighboring examples in representation-space
correspond to more likely input configurations. Indeed, interpolating between
points on a flat manifold should stay on the manifold. Furthermore, when interpolating
between examples of different classes (i.e., different modes), {\bf H3(b)} would
suggest that most of the points in between (on the linear interpolation line)
should correspond to plausible samples, which would not be the case in input space.
In Fig.~\ref{fig:interpol}, we interpolate linearly between neighbors in
representation space and visualize in the input space the interpolated points obtained at
various depths. One can see that interpolating at deeper levels gives visually
more plausible samples.

\subsection{Measuring Mixing by Counting Number of Classes Visited}
\label{sec:mix-time}

We evaluate here the ability of mixing among various classes. We consider sequences
of length 10, 20 or 100 and compute histograms of number of different classes visited
in a sequence, for the two different depths and learners, on TFD. Since classes
typically are in different modes (manifolds), counting how many different classes
are visited in an MCMC run tells us how quickly the chain mixes.
Results in
Fig.~\ref{fig:entropy}(c,f) show that the deeper architectures visit more classes
and the CAE mixes faster than the DBN. 

\subsection{Occupying More Volume Around Data Points}

\begin{figure*}
\begin{center}
\subfigure[TFD]
{\hspace*{-4mm}\includegraphics[width=0.35\textwidth]{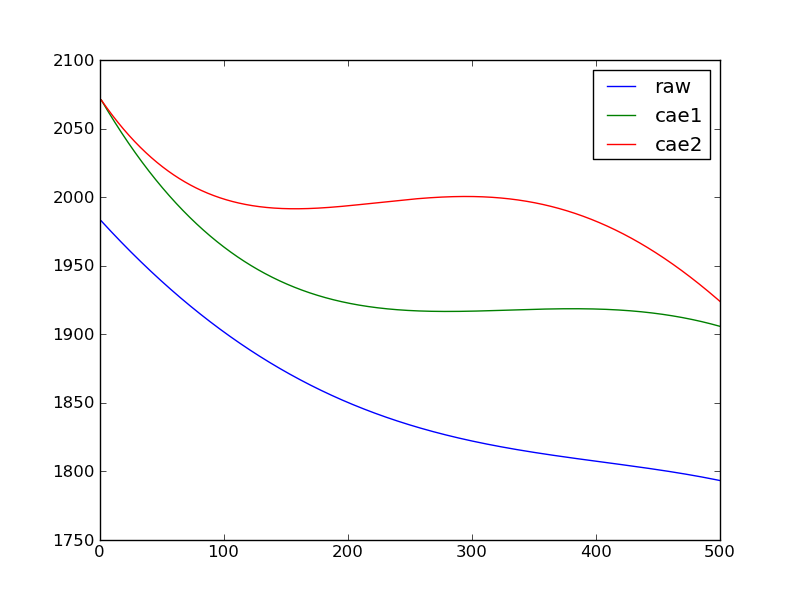}}
\subfigure[MNIST]
{\hspace*{-4mm}\includegraphics[width=0.35\textwidth]{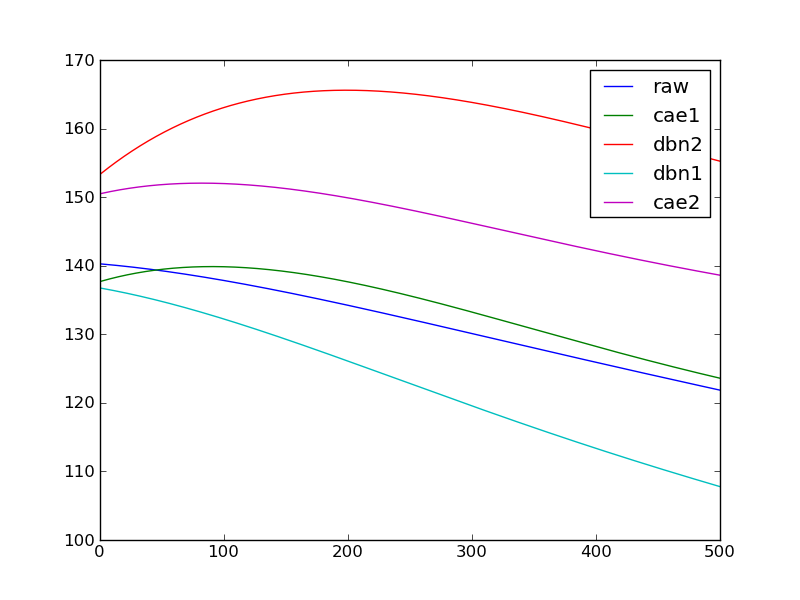}}
\subfigure[Mixing - 10 samples]
{\hspace*{-4mm}\includegraphics[width=0.35\textwidth]{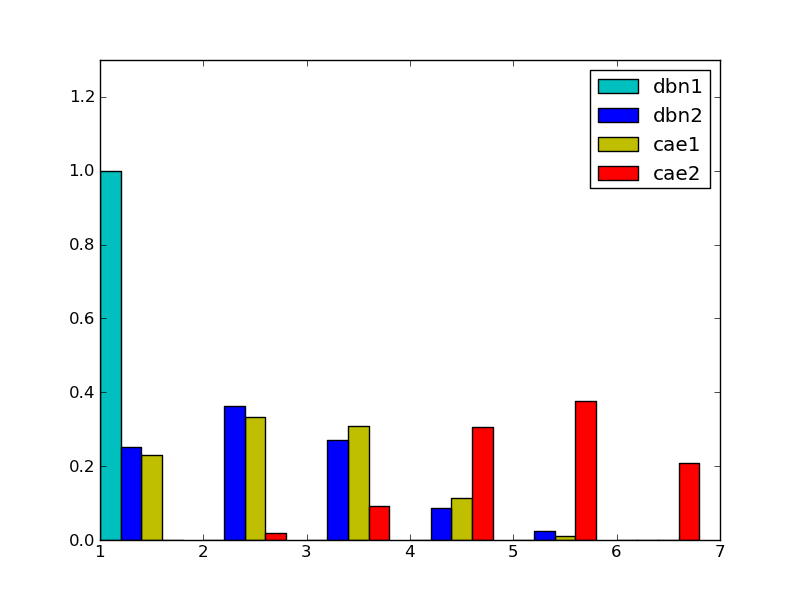}}
\subfigure[TFD]
{\hspace*{-4mm}\includegraphics[width=0.35\textwidth]{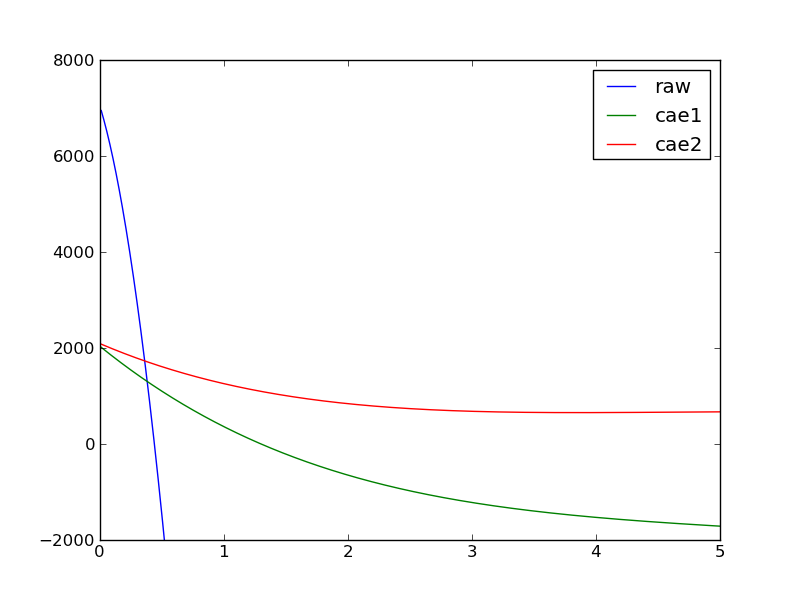}}
\subfigure[MNIST]
{\hspace*{-4mm}\includegraphics[width=0.35\textwidth]{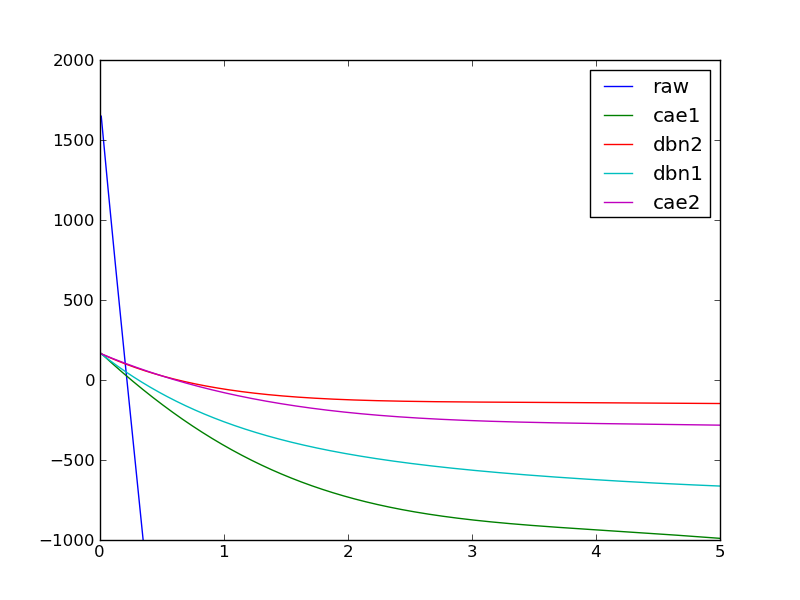}}
\subfigure[Mixing - 20/100 samples]
{\hspace*{-4mm}\includegraphics[width=0.35\textwidth]{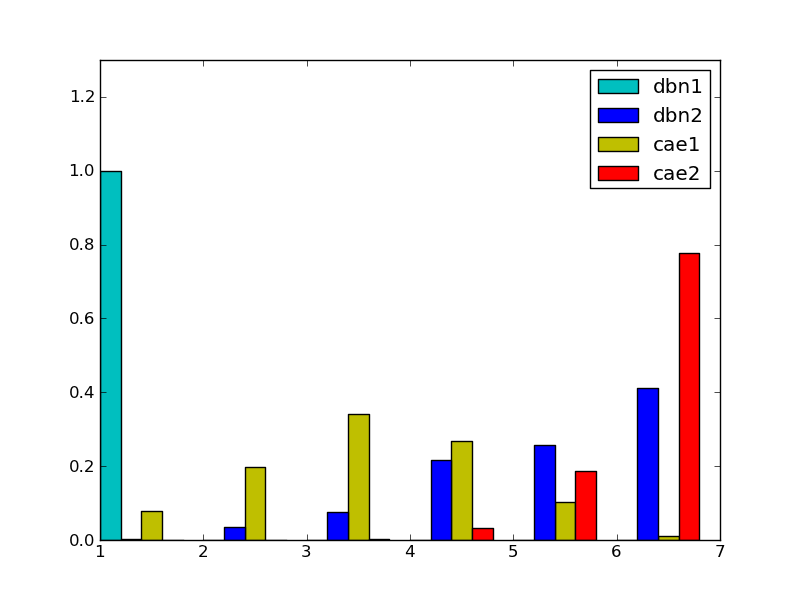}}
\end{center}

\caption{ (a) (b) Local Convex Hull - Log-likelihoods computed w.r.t. linearly
interpolated samples between an example and its k-NNs, for k between 1 and 500. The manifold is unfolded
in deeper levels. (d) (e) Local Convex Ball -
Log-likelihoods of samples generated by adding Gaussian noise to the representation
at different levels ($\sigma\in[0.01,5]$). More volume is taken by good samples on
deeper layers. (c) (f) Mixing Histograms - number of classes visited (x-axis)
over 10 samples (c), 20 samples (f) with CAE, 100 samples (f) with DBN.}
\label{fig:entropy}
\end{figure*}

In these experiments (Fig.~\ref{fig:entropy} (a,b,d,e)) we estimate the quality of samples whose representation is in
the neighborhood of training examples, at different levels of representation.
In the first setting (Fig.~\ref{fig:entropy} (a,b)), the samples are interpolated at the midpoint
between an example and its $k$-th nearest neighbor, with $k$ on the x-axis. 
In the second case (Fig.~\ref{fig:entropy} (d,e)),
isotropic noise is added around an example, with noise standard deviation on the x-axis. 
In both cases, 500 samples are generated
for each data point plotted on the figures, with the y-axis being the log-likelihood
introduced earlier, i.e., estimating the quality of the samples.
We find that on higher-level representations of both the CAE and DBN, a much
larger proportion of the local volume is occupied by likely configurations,
i.e., closer to the input-space manifold near which the actual data-generating
distribution concentrates. Whereas the first experiment shows that this is true
in the convex set between neighbors at different distances (i.e., in the directions
of the manifold), the second shows that this is also true in random directions
(but of course likelihoods are also worse there). The first result therefore
agrees with {\bf H3(a)} (unfolding) and {\bf H3(b)} (volume expansion), while
the second result mostly confirms {\bf H3(b)}.

\subsection{Discriminative Ability vs Volume Expansion}

The proposed hypotheses could arguably correspond to worse discriminative power.
Indeed, if on the higher-level representations the different classes are ``closer''
to each other (making it easier to mix between them), would it not mean that they
are more confusable? We first confirm with the tested models (as a sanity check)
that the deeper level features are conducive to better classification performance,
in spite of their better generative abilities and better mixing.

We train a linear SVM on the concatenation of the raw input with the upper
layers representations.  Results presented in Table~\ref{tab:res-knn-svm} show
that the representation is more linearly separable if one increases the depth of
the architecture and the information added by each layer is helpful for
classification. Also, fine-tuning a MLP initialized with those weights is still
the best way to reach state-of-the-art performances.  

\begin{table*}[ht]
\begin{small}
 \begin{center}
\begin{tabular}{c|c|c||c|c||c|c|}
& \multicolumn{4}{|c||}{\bf Classification} & \multicolumn{2}{|c|}{\bf Log-likelihood}  \\ \hline
& \multicolumn{2}{|c||}{\bf MNIST} & \multicolumn{2}{c||}{\bf TFD} & \multicolumn{1}{|c|}{\bf MNIST} & \multicolumn{1}{c|}{\bf $\,\,\,$TFD $\,\,\,\,$}\\
& SVM &  MLP+ & SVM  & MLP+ & & \\ \hline
raw   & 8.34\%  & -      & $33.48$ {\scriptsize $\pm 2.14$ }\% & - & - & - \\ \hline
CAE-1 & 1.97\%  & 1.14\% & $25.44$ {\scriptsize $ \pm2.45$}\% & $24.12$ {\scriptsize $\pm1.87$} \% & $ 67.69$ {\scriptsize $\pm2.87$} & $591.90$ {\scriptsize $\pm12.27$}\\
CAE-2 & 1.73\%  & 0.81\% & $24.76$ {\scriptsize $\pm2.46$}\% & $23.73$ {\scriptsize $\pm1.62$}\% & $121.17$ {\scriptsize $\pm1.59$} & $2110.09$ {\scriptsize $\pm49.15$}\\ \hline
DBN-1 & 1.62\%  & 1.21\% & $26.85$ {\scriptsize $\pm1.62$}\% & $28.14${\scriptsize $\pm1.40 $}  & $-243.91$ {\scriptsize $\pm54.11$} &  $604$ {\scriptsize $\pm14.67$} \\
DBN-2 & 1.33\%  & 0.99\% & $26.54$ {\scriptsize $\pm1.91$}\% & $27.79$ {\scriptsize $\pm2.34$}& $137.89$ {\scriptsize $\pm2.11$} & $1908.80$ {\scriptsize $\pm65.94$}\\\hline
\end{tabular}

\caption{Left: Classification rates of various classifiers using representations
  learned on the MNIST and TFD datasets.  The DBN 0.99\% error on
  MNIST has been obtained with a 3-layers DBN and the 0.81\% error
  with the Manifold tangent
  Classifier~\citep{Dauphin-et-al-NIPS2011-small} that is based on a
  CAE-2 and discriminant fine-tuning. MLP+ uses discriminant
  fine-tuning. Right: Log-likelihoods from Parzen-Windows density
  estimators based on $10,000$ samples generated by each model. This
  quantitatively confirms that the samples generated from deeper
  levels are of higher quality, in the sense of better covering the
  zones where test examples are found.}
\label{tab:res-knn-svm}

\end{center}
\end{small}
\end{table*}

To explain the good discriminant abilities of the deeper layers (either
when concatenated with lower layers or when fine-tuned discriminatively)
in spite of the better mixing observed, we conjecture the help of a
better disentangling of the underlying factors of variation, and in 
particular of the class factors. This would mean that the manifolds associated 
with different classes are more unfolded (as assumed by {\bf H3(a)}) and possibly
that different hidden units specialize more to specific classes than they
would on lower layers. Hence the unfolding ({\bf H3(a)})
and disentangling ({\bf H1}) hypotheses reconcile better discriminative ability
with expanded volume of good samples ({\bf H3(b)}).

\section{Conclusion}

The following hypotheses were tested: (1) deeper representations can yield better samples
and better mixing; (2) this is due to better disentangling; (3) this is associated with
unfolding of the manifold where data concentrate along with expansion of the volume
good samples take in representation-space. The experimental results were in agreement
with these hypotheses. They showed better samples and better mixing on higher levels,
better samples obtained when interpolating between examples at higher levels, and
better samples obtained when adding isotropic noise at higher levels. We also considered
the potential conflict between the third hypothesis and better discrimination
(confirmed on the models tested) and explained it away as a consequence of the second
hypothesis.

This could be immediate good news for applications requiring to generate MCMC samples:
by transporting the problem to deeper representations, better and faster results could
be obtained. Future work should also investigate the link between better mixing
and the process of training deep learners themselves, when they depend on an MCMC to estimate the log-likelihood
gradient.  One interesting direction is to investigate the relation between parallel tempering and
the better mixing chains obtained from deeper layers.

\bibliography{strings-shorter,ml,aigaion-shorter}
\bibliographystyle{natbib}

\end{document}